\documentclass[letterpaper, 10 pt, conference]{ieeeconf}  

\IEEEoverridecommandlockouts                              

\usepackage{times}
\usepackage[pdftex]{graphicx}
\usepackage{amsmath,amssymb,amsopn,amstext,amsfonts}
\usepackage{cancel}
\usepackage[space]{cite}
\usepackage{pdfsync}
\usepackage{balance}
\usepackage{color}
\usepackage{mathtools}
\usepackage{algpseudocode}
\usepackage[ruled,vlined, linesnumbered]{algorithm2e}
\usepackage{bm}

\usepackage{diagbox}
\usepackage{float}
\usepackage{epstopdf}
\usepackage{pifont}
\usepackage{amsmath}
\usepackage{multirow}
\usepackage{url}
\usepackage{verbatim}
\usepackage[linkcolor=black,citecolor=black,urlcolor=black,colorlinks=true]{hyperref}
\usepackage{booktabs}
\usepackage{graphicx}
\usepackage{subcaption}
\usepackage{threeparttable}  
\usepackage{tikz}
\usepackage{pifont}
\usepackage{amssymb}

\makeatletter
\newcommand*{\rom}[1]{\expandafter\@slowromancap\romannumeral #1@}
\makeatother

\newcommand*{\circled}[1]{\lower.7ex\hbox{\tikz\draw (0pt, 0pt)%
		circle (.5em) node {\makebox[1em][c]{\small #1}};}}

\newtheorem{thm}{Theorem}

\newtheorem{pro}{Problem}
\newtheorem{defi}{Definition}

\title{\LARGE \bf
Air-FAR: Fast and Adaptable Routing for Aerial Navigation in Large-scale Complex Unknown Environments
}

\author{Botao He\textsuperscript{1}, 
	Guofei Chen\textsuperscript{2}, 
        Cornelia Fermuller\textsuperscript{1},
	Yiannis Aloimonos*\textsuperscript{1},
        and Ji Zhang\textsuperscript{2}
	\thanks{1 Perception and Robotics Group, University of Maryland, MD 20742.} 
	\thanks{2 Robotics Institute, Carnegie Mellon University, PA 15213-3890.}
	\thanks{Email: {\tt\small jyaloimo@umd.edu, zhangji@cmu.edu}}
}

\usepackage{amsmath,amssymb,graphicx}
\usepackage{xcolor}

\usepackage{tabularx}


\begin{document}
\vspace{-1.0cm}
\makeatletter
\g@addto@macro\@maketitle{
\begin{figure}[H]
  \setlength{\linewidth}{\textwidth}
  \setlength{\hsize}{\textwidth}
    \centering
    \includegraphics[width=1.0\textwidth]{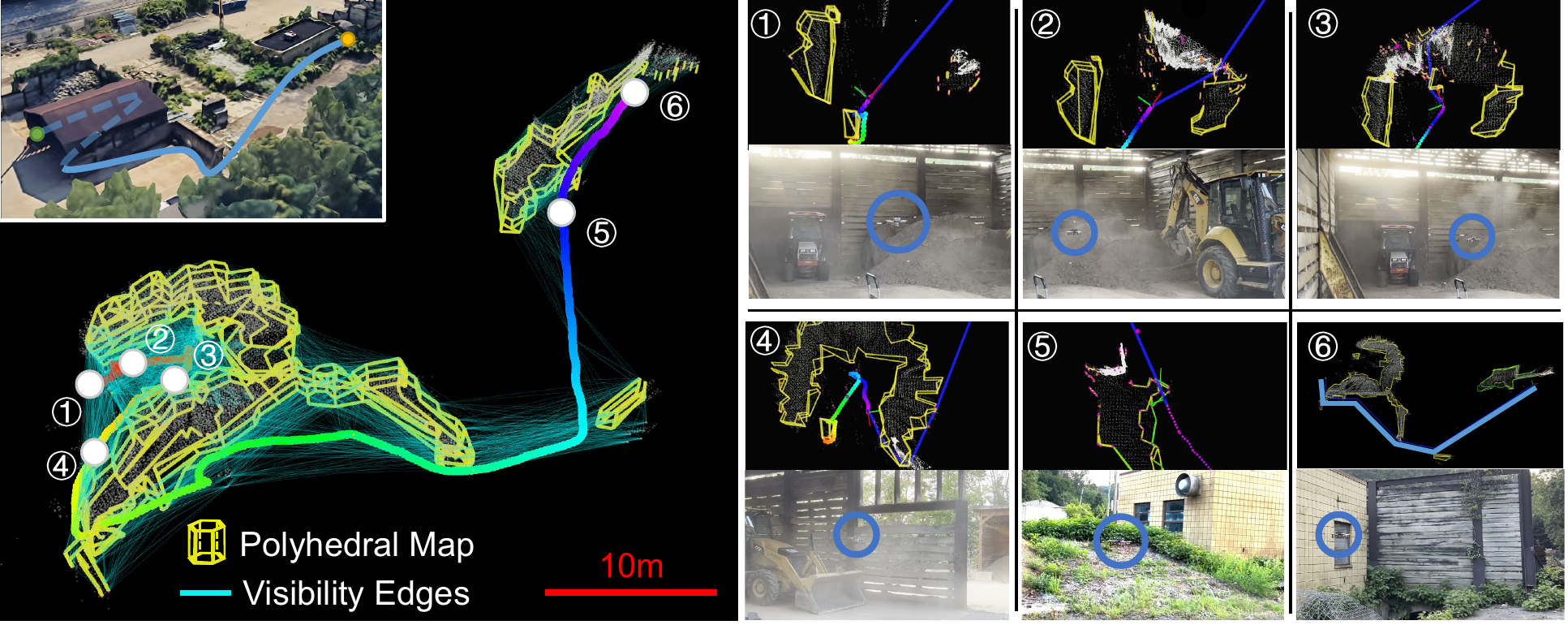}
    \captionsetup{font={small}}
    \caption{Illustration of real-world navigation in a large-scale complex unknown environment with a dead-end. Polyhedral map is marked in yellow and visibility edges are marked in cyan. Robot path is marked in blue and trajectory is marked in rainbow. (1) The robot starts with no prior map. It attempts to approach the goal directly. (2-3) With new sensor observations, the robot incrementally construct the 3D visibility graph (V-graph) in real-time, and adaptively adjust its path  on the updated graph. (4) With the ability to maintain a global graph and perform global path search in real-time, the robot quickly find a path to escape from the dead-end. (5) With the proposed path search and refinement method, our method enables the robot to fly over terrain, which standard V-graph path search methods cannot achieve. (6) After constructing the 3D V-graph, the robot can directly find a near-optimal long-range path within milliseconds.}
    \vspace{-5mm}
    \label{fig:Banner}
\end{figure}
}

\maketitle
\thispagestyle{empty}
\pagestyle{empty}


\null\vspace{-0.6cm}

\begin{abstract}

This paper presents a novel method for real-time 3D navigation in large-scale, complex environments using a hierarchical 3D visibility graph (V-graph). The proposed algorithm addresses the computational challenges of V-graph construction and shortest path search on the graph simultaneously. 
By introducing hierarchical 3D V-graph construction with heuristic visibility update, the 3D V-graph is constructed in $O(K\cdot n^2logn)$ time, which guarantees real-time performance. 
The proposed iterative divide-and-conquer path search method can achieve near-optimal path
solutions within the constraints of real-time operations.
The algorithm ensures efficient 3D V-graph construction and path search.
Extensive simulated and real-world environments validated that our algorithm reduces the travel time by 42\%, achieves up to 24.8\%
higher trajectory efficiency, and runs faster than most benchmarks
by orders of magnitude in complex environments.
The code and developed simulator have been open-sourced to facilitate future research.

\end{abstract}

\section{Introduction}

\setcounter{figure}{1}
\begin{figure}[t]
	\centering
	\includegraphics[width=1.0\linewidth]{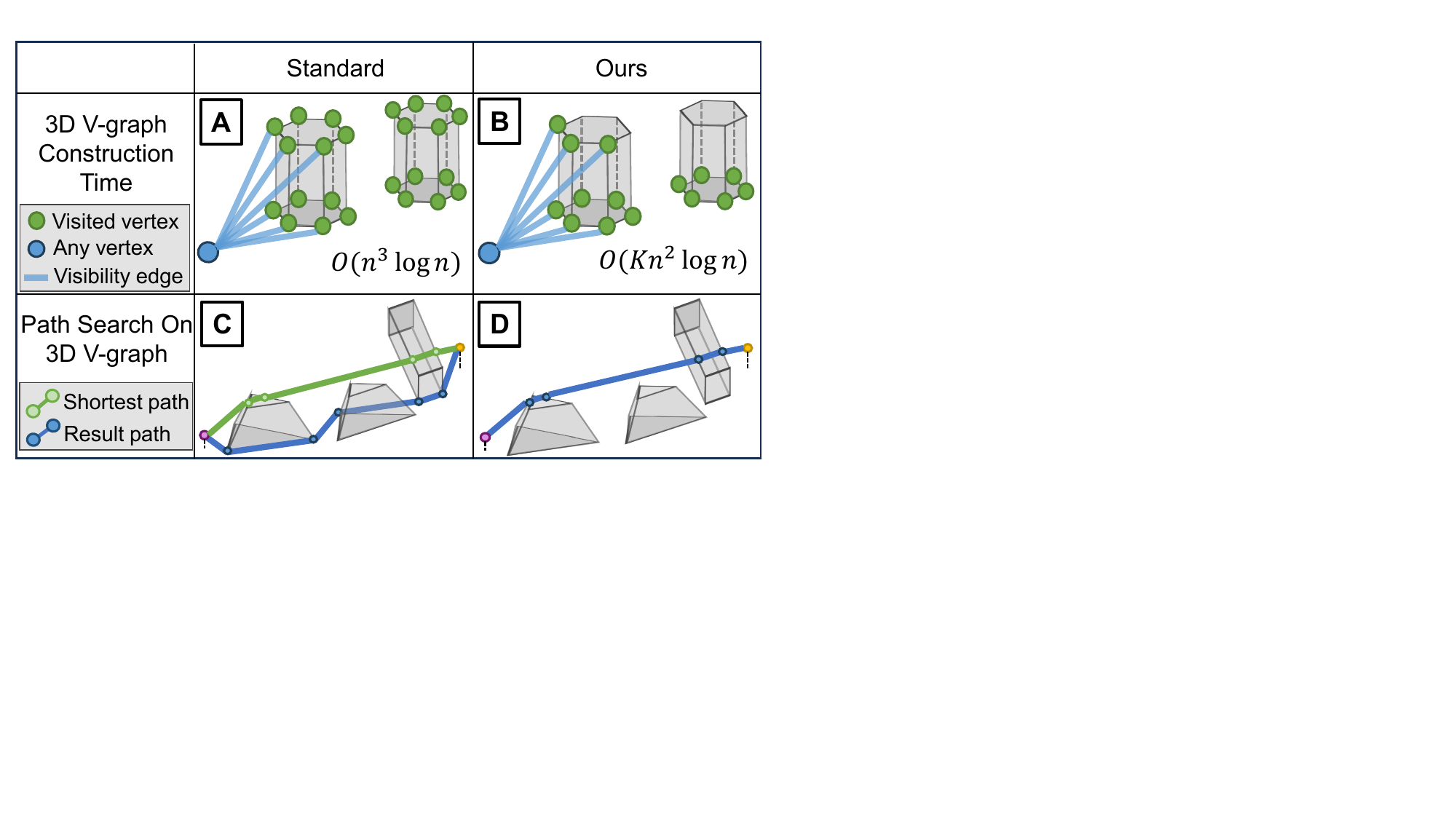}
	\captionsetup{font={small}}
	\caption{ Benefits of the proposed algorithm. 
    }
	\label{fig:benefits}
	\vspace{-0.8cm}
\end{figure}

3D navigation in large-scale complex environments 
remains a challenge. Search-based methods such as A*\cite{hart1968formal}
densely discretize the space and guarantee optimal paths,
but are computationally expensive to propagate on
dense maps. Sampling-based methods, such as RRT*, are
efficient in high-dimensional spaces, but due to their probabilistic sampling mechanism, the performance varies greatly depending on
the environment and takes a long time to find a feasible path in complex environments. Visibility graph(V-graph)-based methods, by
contrast, model the space sparsely by polygonal maps and
connecting only vertices on polygons that are visible to each
other. This sparse representation improves path search speed and memory consumption in orders of magnitude compared to its counterparts. Moreover, it is deterministic, ensuring that a path can always be
found within a bounded time if it exists. In practice, the computation time
remains under 20ms, even for paths exceeding 300 meters.

2D V-graph has proven its efficiency for 2D navigation in cluttered unknown environments due to its sparsity and scalability, however, extending V-graph-based navigation from 2D to 3D is non-trivial. Challenges arise in two aspects: 1) graph construction takes significantly more time because of the increased complexity of three-dimensional spaces, as shown in Fig. \ref{fig:benefits}(A), and 2) motion planning on the graph becomes NP-hard. Specifically, the optimal path cannot be directly calculated by searching vertices alone since it may pass through edges, as illustrated in Fig. \ref{fig:benefits}(C). 

This work addresses both challenges simultaneously. As shown in Fig. \ref{fig:benefits}(B, D), the advantage of our proposed method is its ability to construct the 3D V-graph in real-time and search for a near-optimal path while with real-time guarantee.
Two key ideas underpin the strength of our approach.

The first idea is hierarchical 3D V-graph construction with heuristic visibility update. 
The 3D V-graph is updated hierarchically at each data frame.
While the hierarchical structure and update mechanism aligns with our previous work \cite{yang2022far}, the local-graph construction introduces a novel approach. 
Instead of directly extending the 2D polygonal map into a 3D polyhedral map, which would increase computation by an order of magnitude and thus hinder its real-time performance, we convert sensor data into a layered polygon map and heuristically expand vertical connections between layers. 
These vertical edges transform the layered polygon map into a 3D polyhedral map, with visibility edges connected using similar heuristics. 
The proposed method offers several benefits: 1) heuristic visibility updates decrease computation by an order of magnitude while maintaining good connection density; 2) the layered graph representation is compatible with the 2D V-graph developed in our previous work\cite{yang2022far}, making our algorithm suitable for aerial-ground collaborative navigation.

The second key idea is an iterative divide-and-conquer approach to path searching. The graph's sparsity enables rapid search for an initial path, typically within 10ms for paths over 500 meters. However, the initial path may be low quality, as shown in Fig. \ref{fig:benefits}(C), so sampling nodes on edges is crucial for improvement. Direct sampling on the 3D V-graph is challenging due to the computational cost of visibility updates, limiting the number of points that can be sampled in real-time.
Our method refines the initial path iteratively by trying to connect non-consecutive waypoints of the initial path with the shortest path on the graph. To enhance efficiency, we divide the path into $\log(n)$ subsets for heuristic sampling and re-compute the path after integrating sampled points. This iterative process progressively improves path quality.
The benefits of this approach include: 1) consistently finding an available initial path in real-time, which is crucial for field applications; 2) enabling the robot to fly over obstacles effectively; and 3) achieving near-optimal path solutions within the constraints of real-time operations.

To validate the effectiveness and robustness of our algorithm, we conducted comprehensive tests across 12 simulated large-scale environments with varied complexity and sensor configurations in our developed Autonomy Development Environment. We detail the algorithm's performance in two representative scenarios in this paper.  
We also implemented the algorithm in indoor and outdoor real-world experiments using a custom quadrotor equipped with fully onboard sensing and computing capabilities. 
Extensive experiments reveal that our algorithm outperforms most benchmarks by orders of magnitude, demonstrating enhanced effectiveness in various settings, including indoor, outdoor, simulated, and real-world environments.

We open-sourced the project, including the source code\footnote{Air-FAR: \tt \href{https://github.com/Bottle101/Air-FAR}{github.com/Bottle101/Air-FAR}} and the Autonomy Development Environment\footnote{Autonomy Dev. Env.: \tt \href{https://github.com/Bottle101/Air-FAR}{\\ www.far-planner.com/development-environment}}, to facilitate further research.

\section{Related Work}
\subsection{Search- or Sampling-based Path Search for 3D Navigation in unknown environments}
Search-based planners like Dijkstra's \cite{dijkstra1959note} and A* \cite{hart1968formal} discretize space densely and guarantee optimal paths but are computationally intensive for large, dense maps. Sampling-based planners, such as RRT \cite{LaValle1998RapidlyexploringRT} and RRT* \cite{karaman2011sampling}, handle high-dimensional spaces well but lack guaranteed convergence within a time limit and struggle with large, complex environments or dynamic global map updates.

To mitigate these challenges, some approaches limit the map size to a smaller local area without maintaining global maps. For example, Zhou et al. \cite{zhou2019robust}\cite{zhou2021raptor} used Hybrid A* \cite{dolgov2008practical} in a $10m \times 10m \times 3m$ local map, while Ye et al. \cite{ye2020tgk} applied Kinodynamic RRT* \cite{webb2013kinodynamic} within a similar range. While these methods achieve real-time performance, they are prone to local minima, potentially trapping the robot if dead-ends extend beyond the map's boundaries.

Our method, using a sparse V-graph, significantly reduces time and memory complexity compared to search-based algorithms. Our approach deterministically finds a path within a 20ms time bound for distances over 300 meters. By maintaining an incrementally updated global map, it is fundamentally robust to the local minima.

\subsection{V-graph-based Navigation}
The V-graph has long been studied \cite{lozano1979algorithm}, but its application in navigation is rare due to high computational costs \cite{kitzinger2003visibility}. Recently, Yang et al.\cite{yang2022far} introduced a hierarchical V-graph to reduce construction time, enabling real-time 2D navigation in cluttered environments. However, 3D V-graph navigation remains a challenge. Bygi et al. \cite{bygi20073d} proposed an algorithm that constructs the 3D V-graph in \(O(n^3 \log n)\) time, Yang et al. \cite{yang2022far} extended their 2D V-graph to 3D using layered polygons, both of them cannot satisfy real-time requirements for navigation in complex environments. 
Furthermore, shortest path search on the 3D V-graph is non-trivial, as it is an NP-hard problem \cite{jiang1993finding}. \cite{you20193d} proposed an optimization-based real-time path planning and control algorithm for 3D V-graphs, but it sacrifices planning horizon for speed, limiting its applicability for long-range navigation tasks.

Our method address both challenges simultaneously. We proposed our heuristic 3D V-graph construction algorithm to build and update the 3D V-graph in real-time. The proposed iterative divide-and-conquer method allows our algorithm search for a near-optimal path while with real-time guarantee. 

\begin{figure*}[t!]
	\centering
	\includegraphics[width=0.8\textwidth]{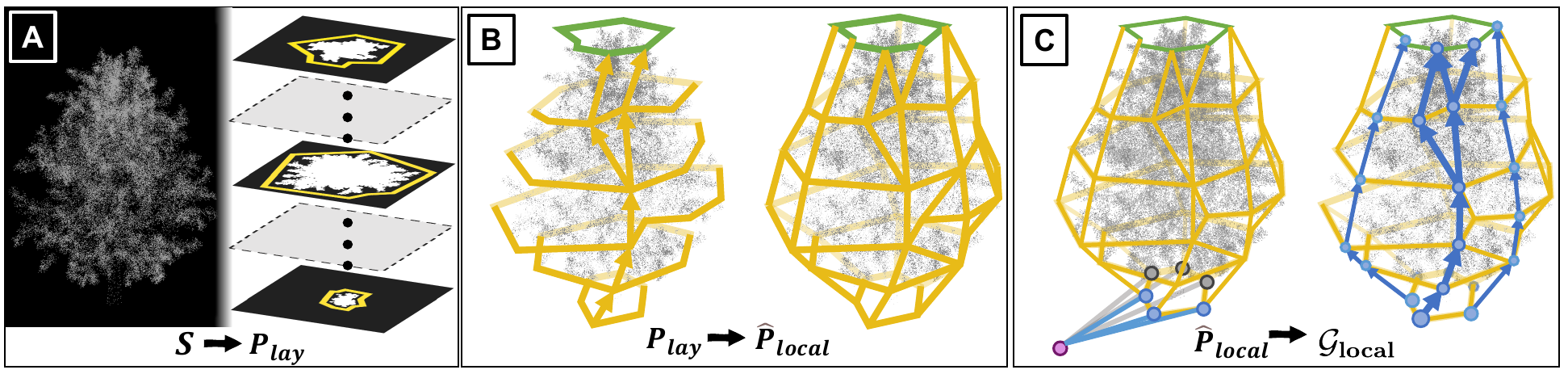}
	\captionsetup{font={small}}
	\caption{
		Illustration of the steps to heuristically construct the 3D V-graph.
	}
	\label{fig:buildgraph}
	\vspace{-0.5cm}
\end{figure*}

\section{Problem Definition}



        
Our problem is divided into two sub-problems: 3D V-graph incremental construction and explorative-optimal path search on the 3D V-graph.

For the first sub-problem, define $\textit{Q} \subset \mathbb{R}^3$ as the workspace for the robot to navigate. Let $\mathcal{S} \subset \textit{Q}$ be the perceived sensor data. We propose a new 3D visibility graph (V-graph) representation, denoted as $\mathcal{G} \subset \textit{Q}$, constructed from $\mathcal{S}$.

\begin{pro}
\vspace{0.05in}
    During navigation in unknown environments, given $\mathcal{S}$, fit the point cloud with polyhedra and incrementally construct the global 3D V-graph in real time.
\vspace{0.05in}
\end{pro}

Problem 1 is solved in two steps. First, we hierarchically separate the $\mathcal{G}$ into two layers: local layer $\mathcal{G}_{local}$ and global layer $\mathcal{G}_{global}$, so $\mathcal{G} = \{ \mathcal{G}^i_{local} \subset \mathcal{Q}, \mathcal{G}^i_{global} \subset \mathcal{Q} | i \in \mathbb{Z}^+\}$. At each sensor frame, we only construct $\mathcal{G}_{local}$ from $\mathcal{S}$ and merge into $\mathcal{G}_{global}$. Second, we build the local polyhedral map using layered polygons and heuristically assess visibility between vertices to balance connection density and computational efficiency.

For the second sub-problem, we first define the concept \textit{explorative-optimal}. 
Since the global optimal path cannot be directly computed in an unknown environment, the current optimal path can only be searched on the currently available map during navigation or exploration. Thus, \textit{explorative-optimal} is defined as follows:

\begin{defi}
\vspace{0.05in}
    A path is called \textit{explorative-optimal} if and only if it satisfies certain optimality criteria under the current environmental observations.
\vspace{0.05in}
\end{defi}

Then, we define a path that has a set of waypoints $\mathbf{P} = \{\mathbf{p}_i \in \mathcal{Q} | i \in \mathbb{Z}^+\}$.
The problem is then defined as an optimal path search problem on 3D V-graph:

\begin{pro}
\vspace{0.05in}
    Given the current $\mathcal{G}$, robot position $\mathbf{p}_{robot} \in \textit{Q}$ and goal point $\mathbf{p}_{goal} \in \textit{Q}$, find the explorative-optimal path $\mathbf{P}^*$ between $\mathbf{p}_{robot}$ and $\mathbf{p}_{goal}$ on $\mathcal{G}$.
\vspace{0.05in}
\end{pro}

Problem 2 is solved repetitively in each planning cycle. 
During navigation, we re-plan the path on the updated $\mathcal{G}$ until arriving at $\mathbf{p}_{goal}$. We propose our iterative divide-and-conquer path search method to achieve an asymptotic explorative-optimal path search, which also has probabilistic guarantees of completeness and optimality. We experimentally validated that our algorithm can achieve near-optimal within 2 iterations under the promise of real-time performance.

\section{3D V-graph Construction and Update}
\subsection{Polyhedron Extraction}
\label{sec:poly_extraction}
The 3D V-graph uses polyhedra to represent 3D obstacles. The polyhedra map proposed in this paper, denoted as $\hat{\mathcal{P}} = \{ \hat{\mathcal{P}}^i_{local} \subset \mathcal{Q}, \hat{\mathcal{P}}^i_{global} \subset \mathcal{Q} | i \in \mathbb{Z}^+\}$, is derived from layered polygon map, which is denoted as $\mathcal{P}_{lay} = \{\mathcal{P}^{i}_{lay} \subset \mathcal{Q} | i \in \mathbb{Z}^+\}$. 
As illustrated in Fig. \ref{fig:buildgraph}(A), to calculate $\mathcal{P}_{lay}$, we first slice the 3D sensor data $\mathcal{S}$ into multiple pieces based on a given resolution, inflate them based on the robot dimension, and register them to 2-D image planes. 
For each slice of data, we extract enclosed polygons using the methods in \cite{suzuki1985topological} and \cite{douglas1973algorithms}, this step is developed from our previous work \cite{yang2022far}. 

The next step involves connecting the vertical contours. As shown in Fig. \ref{fig:buildgraph}(B) and Alg. \ref{alg:graphCon}, for each vertex $\mathbf{v} \in \{\mathcal{P}^{1}_{lay}, ..., \mathcal{P}^{i-1}_{lay}\}$, we search for its K nearest neighbors on the layer above within a radius and connect them to form vertical contour connections. This process enables the construction of a polyhedral map that represents 3D obstacles. The visibility check algorithm further utilizes the vertical contours to enhance its efficiency.
Note that the polyhedra we construct do not need to be fully enclosed.

\subsection{Local 3D V-graph Construction}
\label{sec:graph_construction}
The local 3D V-graph is built in real-time at each sensor frame. To balance between efficiency and connection density, we introduce our heuristic 3D V-graph construction algorithm. 
As shown in Fig. \ref{fig:buildgraph}(C) and the second part of the Alg. \ref{alg:graphCon}, we first check visibility for all same-layer vertices, this step has the time complexity of $O(n^2logn)$ and was proven to promise real-time in \cite{yang2022far}. 
Instead of directly checking visibility for inter-layer vertices, which would result in a time complexity of $O(n^3 \log n)$ \cite{bygi20073d} and is impractical for real-time computation, we heuristically check the visibility for all vertically connected vertices of each same-layer visible vertex. The rationale behind this approach is that if part of an object is already visible, it is more likely to be in the foreground, making other parts of the same polyhedron more likely to be visible as well.
In this way, the total time complexity can be reduced to $O(K \cdot n^2\log{n})$. The notation and analysis are detailed in Sect. \ref{sec:graph_analyse}. 
We enhance graph connectivity and mitigate sensor noise by randomly sampling a few vertices $\mathcal{V}_{sample}$ during each planning cycle, typically fewer than five points. The time required for this step is minimal.
Through extensive evaluation, we found that our algorithm can be executed in real-time while maintaining good graph connectivity. 

\setlength{\textfloatsep}{0pt}
\begin{algorithm}[t]
\SetAlgoLined
\LinesNumbered
\SetKwInOut{Input}{Input}
\SetKwInOut{Output}{Output}
\Input{Layered polygon map: $\mathcal{P}_{lay}$}
\Output{Local V-graph: $\mathcal{G}_{local}$}

\Comment{Polyhedron map construction}

Add $\mathcal{P}^{i}_{lay}$ to $\mathcal{P}_{top}$

\For{each vertex $ \mathbf{v} \in \{\mathcal{P}^{1}_{lay}, ..., \mathcal{P}^{i-1}_{lay}\}$}{
$\mathcal{V}_{vert} \leftarrow KNNRadiusSearch(\mathcal{P}^{k+1}_{lay})$\\
Add $\mathcal{V}_{vert}$ as vertical contour connections to $\mathbf{v}$
}
Add $\mathcal{P}_{lay} \cup \mathcal{V}_{vert}$ to $\hat{\mathcal{P}}_{local}$

\Comment{Connect visibility edges}

\For{each vertex $ \mathbf{v} \in \mathcal{P}_{lay}$}{
Add $CheckVisibility(\mathbf{v}, \mathcal{P}^{i}_{lay})$ to $\mathcal{V}_{visible}$\\
let $Q$ be a queue, 
$Q.enqueue(\mathcal{V}_{visible})$\\
\While {$Q$ is not empty}{
$\mathbf{v}_{vis} \leftarrow Q.dequeue()$\\
$\mathcal{V}_{vert} \leftarrow getVerticalConnections(\mathbf{v}_{vis})$\\
Add $CheckVisibility(\mathbf{v}, \mathcal{V}_{vert})$ to $\mathcal{V}_{visible}$\\
$Q.enqueue(\mathcal{V}_{vert})$\\
}
Add $\mathcal{V}_{visible}$ to $\mathcal{G}_{local}$ 
}

\uIf{ Within time budget}{
$\mathcal{V}_{sample} \leftarrow$ Sample $N$ points on $\mathcal{G}_{local}$\\
Check visibility and add $\mathcal{V}_{sample}$ to $\mathcal{G}_{local}$
}

\Return $\mathcal{G}_{local}$\;

\caption{Local 3D V-graph Construction}\label{alg:graphCon}
\end{algorithm}

\subsection{Two-layer Graph Update}
\label{sec:graph_update}

The graph update method aligns with the approach outlined in our previous work \cite{yang2022far, he2024interactive}. with details available therein. Briefly, during each planning cycle, we compare $\mathcal{G}_{local}$ with $\mathcal{G}_{global}$. For each vertex in $\mathcal{G}_{local}$, we update its position in $\mathcal{G}_{gloal}$ if it has a corresponding vertex there; if not, we introduce it as a new vertex. Conversely, if a vertex exists in $\mathcal{G}_{global}$ but is absent in the corresponding location of $\mathcal{G}_{local}$, we classify it as a disappeared vertex. It will be removed from $\mathcal{G}_{global}$ if it remains absent for several frames.

\subsection{Computational Complexity Analysis}
\label{sec:graph_analyse}
Assume vertices are evenly distributed among layers, and each layer has vertex number $n_{l}$. The relationship between $n_{l}$ and total number of vertices $n$ can be expressed as $n_{l} = \frac{n}{m}$, where $m$ is the number of layers.

\begin{thm}
\vspace{0.05in}
  Each vertex takes at most $O(k \cdot n\log{n})$ time to update its visibility in $\hat{\mathcal{P}}_{local}$.
\vspace{0.05in}
\end{thm}

It is proven that one vertex takes $O(n\log{n})$ time to connect its 2D visibility edges with other same-layer vertices. 
In the 3D case, for any inter-layer vertex pair, a visibility check needs to be performed on all layers between them. For example, for vertex pair $\langle \mathbf{v}_1, \mathbf{v}_3 \rangle$, where the index indicates their layer id, we not only need to do an intersection check on layer 1, but also need to check layer 2 since we do not want the robot to collide with intermediate layers too.
Therefore, assuming each vertex has $k$ vertical contour connections, and all $k$ connections are evenly distributed among $m$ layers, the time consumption to check vertical contour connections is:

\begin{equation}
    \sum_{i=1}^{m}i \cdot \frac{k}{m}n_l\log{n_l} = \frac{k(m+1)}{4}n_l\log{n_l}.
\end{equation}

Therefore, the time consumption to update the visibility is 

\begin{equation}
\begin{aligned}
    \mathcal{T}_{single} &= n_l\log{n_l} + \frac{k(m+1)}{4}n_l\log{n_l}\\
    &= (\frac{1}{m} + \frac{k}{4} \cdot \frac{m+1}{m})n\log{n}
\end{aligned}
\label{eq:singleVertexUpdate}
\end{equation}
Because $m$ is constant and the time complexity changes linearly with $k$, Eq. \ref{eq:singleVertexUpdate} can be expressed as $O(k \cdot n\log{n})$.

\begin{thm}
\vspace{0.05in}
  Time complexity of Alg. \ref{alg:graphCon} is $O(K \cdot n^2\log{n})$.
\vspace{0.05in}
\end{thm}

{\it~~Proof}: Define the total vertex number versus the visible vertex number for a given vertex is $\lambda$. In this way, For a single layer of the graph, we need to perform visibility check $n_l$ times for same-layer vertices and $\frac{n_l}{\lambda}$ times for inter-layer vertices. The time consumption to update the visibility edges of the whole V-graph is 
\begin{equation}
\begin{aligned}
    \mathcal{T} = (\frac{1}{m} + K \cdot \frac{m+1}{m})n^2\log{n}, 
\end{aligned}
\label{eq:allVertexUpdate}
\end{equation}
where $K = \frac{k}{4 \lambda}$.

Another part of Alg. \ref{alg:graphCon} is the KNN search, taking $O{n\log{n}}$ time to construct a KD-tree \cite{wald2006building} and another $O{n\log{n}}$ time to query all vertices, which is ignitable compared with $\mathcal{T}$.
Therefore, the final time complexity is $O(K \cdot n^2\log{n})$. 
In practice, the $\lambda$ is around 10, making the algorithm efficient for real-time computing.





\begin{figure}[t!]
	\centering
	\includegraphics[width=0.95\linewidth]{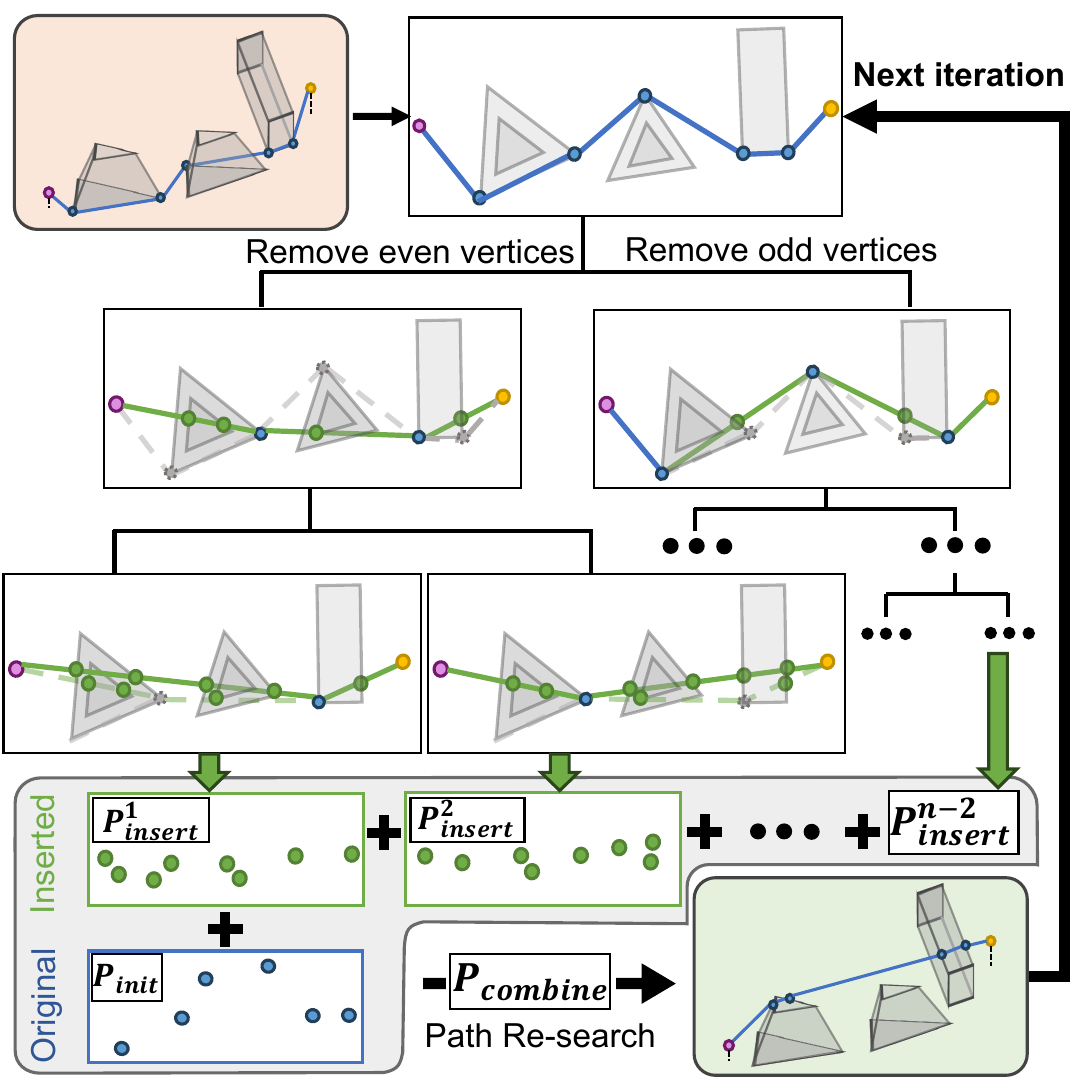}
	\captionsetup{font={small}}
	\caption{
		Illustration for the proposed iterative divide-and-conquer path search algorithm. 
	}
	\label{fig:pathsearch}
	\vspace{-0.3cm}
\end{figure}

\section{Iterative Divide-and-Conquer Path Search}

After constructing the 3D V-graph, searching for an explorative-optimal path within the \textbf{vertex domain} becomes straightforward. However, as previously discussed, the optimal waypoint might be located on an edge. To compute the explorative-optimal path within the \textbf{graph domain}, we introduce our iterative divide-and-conquer path search algorithm, which can search for near explorative-optimal paths under the promise of real-time.

Define $\mathbf{P}_{init} = \{\mathbf{p}_{init}^i \in \mathcal{Q} | i \in \mathbb{Z}^+\}$ as the initial path search result on vertex-domain.
As shown in Fig. \ref{fig:pathsearch} and Alg. \ref{alg:search}, given $\mathbf{P}_{init}$, we decompose it into two two sub-sets $\{\mathbf{P}_{odd}, \mathbf{P}_{even}\} \subset \mathbf{P}_{init}$. 
We re-connect consequent waypoints for each sub-set as new path segments and calculate its intersection with the graph. The intersection points $\mathbf{P}_{insert}$, marked as green in Fig. \ref{fig:pathsearch}, are saved, and their visibility with other vertices is calculated.
We only check intersections with the top layer contours because, as discussed in the Introduction, the more optimal paths will likely fly over obstacles. The problem is repeatedly divided into two sub-problems until only one waypoint from $\mathbf{P}_{init}$ remains. 
After expanding all sub-problems, we merge all inserted waypoints with the initial waypoints and re-search the path to compute a refined path. This refined path is then used as the input for the next iteration. We iteratively solve the problem until convergence is achieved or the time limit is reached.

\subsection{Admissibility, Completeness and Optimality}

Every time we check intersection between two non-consecutive waypoints $\langle \mathbf{p}_{init}^i, \mathbf{p}_{init}^j \rangle$ and $\hat{\mathcal{P}}_{global}$, we try to connect $\mathbf{p}_{init}^i$ and $\mathbf{p}_{init}^j$ with shortest path on the graph, therefore the heuristic is admissible. 
Benefited from the direct sampling part in Alg. \ref{alg:graphCon}, the completeness and asymptotic explorative optimal can be guaranteed in principle. However, in practical applications, optimality and completeness are often compromised to guarantee real-time performance. Extensive experiments validated that the path obtained after one or two iterations is near-optimal and sufficiently effective for the robot to execute.

\setlength{\textfloatsep}{0pt}
\begin{algorithm}[t!]
\SetAlgoLined
\LinesNumbered
\SetKwInOut{Input}{Input}
\SetKwInOut{Output}{Output}
\Input{Initial Path: $\mathbf{P}_{init}$}
\Output{Refined Path: $\mathbf{P}_{refine}$}

let $Q$ be a queue, $Q.enqueue(\mathbf{P}_{init})$ \Comment{Divide} \\
\While {$Q$ is not empty \textbf{and} within time budget}{
$\mathbf{P} \leftarrow Q.dequeue()$\\
$\mathbf{P}_{odd} \leftarrow RemoveEvenNodes(\mathbf{P})$\\
$\mathbf{P}_{even} \leftarrow RemoveOddNodes(\mathbf{P})$\\

\For{each waypoint pair $ \langle \mathbf{p}_i, \mathbf{p}_{i+1} \rangle \in \mathbf{P}_{odd}$}{
$\mathbf{P}_{intersect} \leftarrow Intersection(\langle \mathbf{p}_i, \mathbf{p}_{i+1} \rangle, \mathcal{P}_{top})$\\
$CheckVisibility(\mathbf{P}_{intersect}, \hat{\mathcal{P}}_{global})$\\
Add $\mathbf{P}_{intersect}$ to $\mathbf{P}_{insert}$
}

\uIf{ The size of $\mathbf{P}_{odd} \backslash \{\mathbf{p}_{robot}, \mathbf{p}_{goal}\} > 1$}{
$Q.enqueue(\mathbf{P}_{odd})$
}
\Comment{Apply the same operation for $\mathbf{P}_{even}$}
}

$\mathbf{P}_{combine} \leftarrow Merge(\mathbf{P}_{insert}, \mathbf{P}_{init})$ \Comment{Conquer} \\

$\mathbf{P}_{refine} \leftarrow$ path re-search on $\mathbf{P}_{combine}$ using \cite{dijkstra1959note}\\

Use $\mathbf{P}_{refine}$ for the next iteration

\caption{One Iteration of the Path Search}\label{alg:search}
\end{algorithm}

\section{Experiments}
\subsection{Experiment Setup}
\subsubsection{Simulation}

\begin{figure}[t!]
	\centering
	\includegraphics[width=1.0\linewidth]{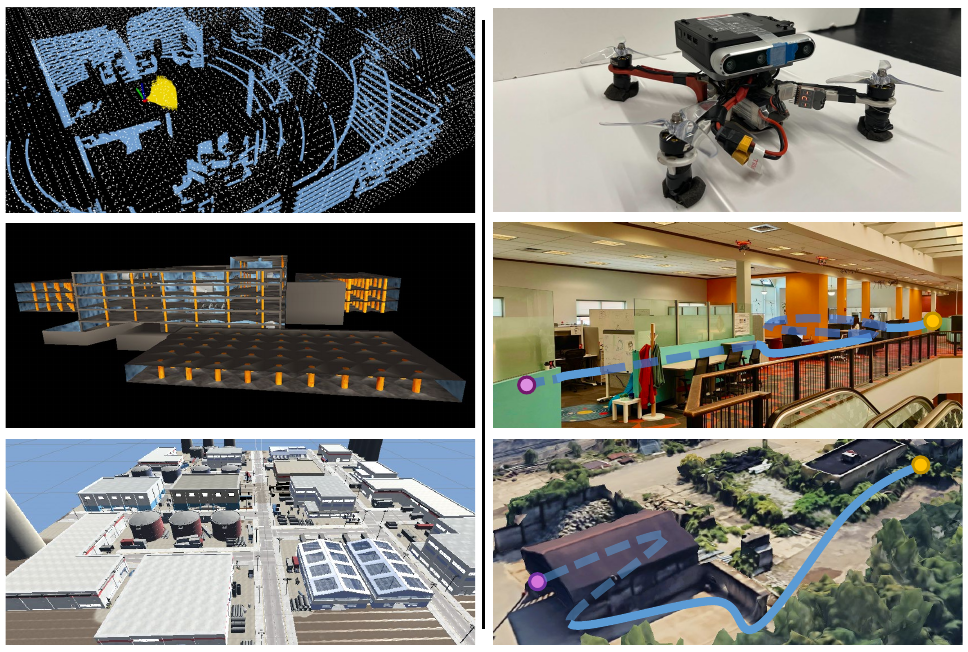}
	\captionsetup{font={small}}
	\caption{
		Experiment setup. The left column shows the simulated drone with a Lidar, overview of the garage and the factory. The right column shows the customized quadroter with a depth camera, overview of the lab and the outdoor space.
	}
	\label{fig:env}
	\vspace{-0.8cm}
\end{figure}

Our Autonomy Development Environment for this project features 29 multi-scale scenes with varied complexity, supporting both ground and aerial navigation. 
We test our algorithm in two complex environments depicted in Fig. \ref{fig:env}(A): a $140 \times 130m$ indoor garage and a $300 \times 300m$ outdoor factory. The robot utilizes LiDar for navigation. The framework operates on a laptop with an i7-12700H CPU, updating the 3D V-graph at 7.5Hz and conducting path searches with each update. We set the spatial resolution at $0.15m$, and the local layer covers a $60 \times 60m$ area with the vehicle in the center.

\subsubsection{Real-world}


We validated our algorithm in real-world settings using a custom quadrotor equipped with a Realsense D455 depth camera and an Intel i5-1135G7 processor. The first test environment was a complex 20×15m indoor area, used for the 3D V-graph testing construction and update. The second environment, a 45×20m outdoor space with a dead-end, served for comprehensive system-level testing.

\begin{figure*}[!t]
    \centering
    \begin{minipage}{0.43\textwidth}
        \centering
        \includegraphics[width=0.9\textwidth]{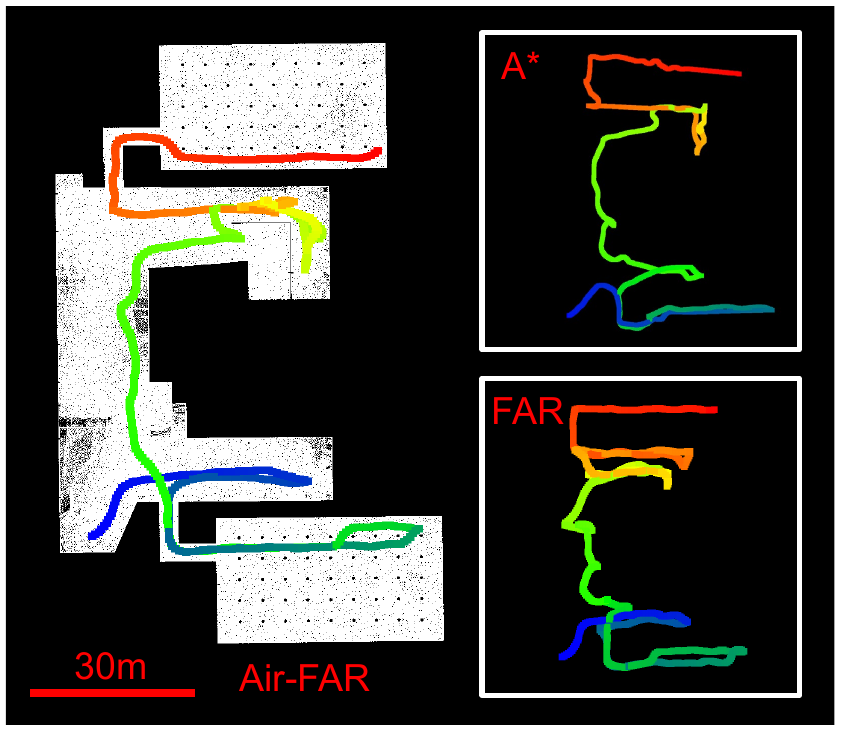} 
        \captionsetup{font={small}}
        \caption{The resulting map and trajectories of system-level experiment in Garage.}
        \label{fig:garage}
    \end{minipage}\hfill
    \begin{minipage}{0.43\textwidth}
        \centering
        \includegraphics[width=0.9\textwidth]{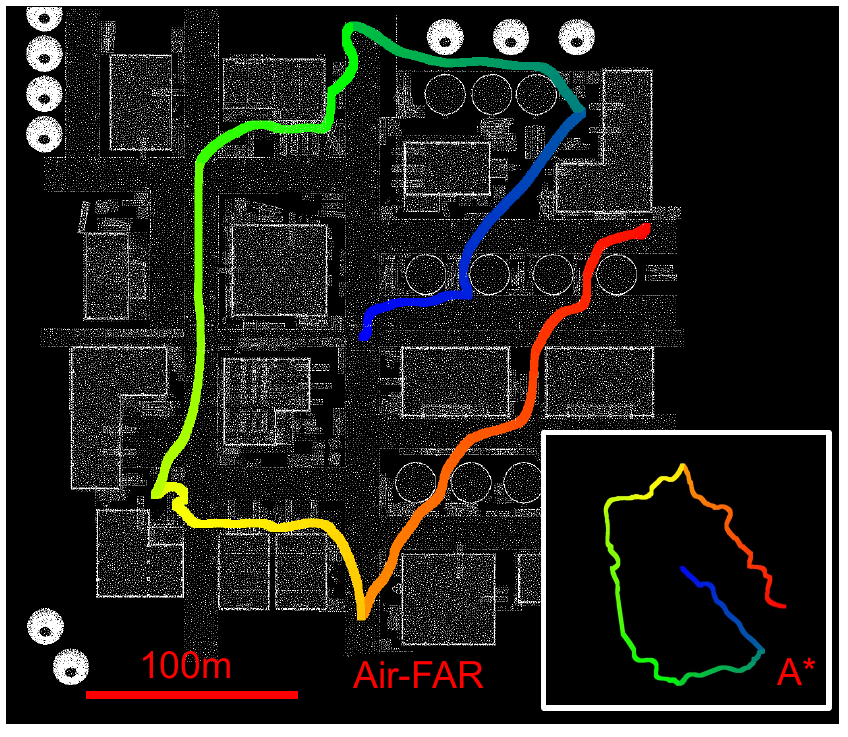} 
        \captionsetup{font={small}}
        \caption{The resulting map and trajectories of system-level experiment in Factory.}
        \label{fig:factory}
    \end{minipage}
\vspace{-0.7cm}
\end{figure*}


\subsection{3D V-graph Construction, Computational and Path Efficiency Comparison}
\label{sec:efficiency}
To comprehensively demonstrate the efficiency and effectiveness of our algorithm, we compared graph construction time, path search time, and path quality across various environments. We evaluated our method against several benchmarks: the search method \textbf{A*}, the sampling-based \textbf{RRT*} and \textbf{BIT*} \cite{qureshi2015intelligent}, and \textbf{FAR} \cite{yang2022far}. \textbf{BIT*} represents the state-of-the-art in sampling-based methods. The ground truth (GT) path is provided by \textbf{A*}. 
For RRT* and BIT*, we noted the refined path time when it was within 1.05 times the GT path length.

As shown in Table \ref{tab:time}, the proposed method can update the 3D V-graph in real-time,
and outperforms all other methods by orders of magnitude in search time for both the initial path and refining to near optimal. 
As shown in Table \ref{tab:pathlength}, though cannot promise optimal, our algorithm can provide a near-optimal solution in real-time, which is crucial for navigation in unknown environments.

\begin{table}[t!]
\centering
\caption{Average Search Time in [ms]}
\label{tab:time}
\resizebox{\linewidth}{!}{%
\begin{tabular}{|c|cc|cc|}
\hline
Env                                                         & \multicolumn{2}{c|}{Garage}      & \multicolumn{2}{c|}{Factory}        \\ \hline
\begin{tabular}[c]{@{}c@{}}Path Len. (m)\end{tabular}  & \multicolumn{2}{c|}{110}       & \multicolumn{2}{c|}{323}            \\ \hline
\begin{tabular}[c]{@{}c@{}}Search \\ Time (ms)\end{tabular} & Initial Path   & Refined         & Initial Path      & Refined         \\ \hline
A*                                                          & 272.7         & -               & 3.5e4            & -               \\ \hline
RRT*                                                        & \textgreater{}1e4              & -           & 6.4e3 & 3.0e4      \\ \hline
BIT*                                                        & 40.2              & 275.2             & 4.6e3                 & 9.0e3          \\ \hline
FAR                                                         & 12.8         & -               & -        & -      \\ \hline
Ours                                                        & \textbf{7.1} & \textbf{18.9} & \textbf{15.5}   & \textbf{61.8} \\ \hline
\end{tabular}%
}
\vspace{-1.0cm}
\end{table}

\begin{table}[b]
\centering
\caption{Average 3D V-graph Update Time in [ms]}
\label{tab:graphupdatetime}
\resizebox{0.7\linewidth}{!}{%
\begin{tabular}{|c|c|c|c|}
\hline
Env                          & Garage                     & Factory                    & Lab                           \\ \hline
\multicolumn{1}{|l|}{Sensor} & \multicolumn{1}{l|}{Lidar} & \multicolumn{1}{l|}{Lidar} & \multicolumn{1}{l|}{DepthCam} \\ \hline
FAR                          & 406.4                      & 4651.5                     & 382.3                         \\ \hline
Ours                         & \textbf{81.6}              & \textbf{153.6}             & \textbf{112.7}                \\ \hline
\end{tabular}%
}
\end{table}

\begin{table}[t]
\centering
\caption{Average Path Quality in [\%]}
\label{tab:pathlength}
\resizebox{\linewidth}{!}{%
\begin{tabular}{|c|cc|cc|}
\hline
Env                                                         & \multicolumn{2}{c|}{Garage} & \multicolumn{2}{c|}{Factory} \\ \hline
\begin{tabular}[c]{@{}c@{}}Path Len. (m)\end{tabular}  & \multicolumn{2}{c|}{110}  & \multicolumn{2}{c|}{323}     \\ \hline
\begin{tabular}[c]{@{}c@{}}Path \\ Quality (\%)\end{tabular} & Initial Path  & Refined     & Initial Path  & Refined      \\ \hline
A*                                                          & \textbf{100}   & \textbf{100} & \textbf{100}   & \textbf{100}  \\ \hline
RRT*                                                        & -          & -        & 75.5             & 76.1   \\ \hline
BIT*                                                        & 50.2          & 96.9        & 88.0             & 95.3         \\ \hline
FAR                                                         & 98.1          & 98.1        & -    & -   \\ \hline
Ours                                                        & 97.3          & 97.3        & 95.8          & 97.6         \\ \hline
\end{tabular}%
}
\vspace{-0.55cm}
\end{table}

\subsection{System-level Comparison}
To illustrate the reliability of our algorithm in field application scenarios, we compare the systematic navigation performance with A* based Ego-Planner\cite{zhou2021ego} and FAR-Planner\cite{yang2022far}. Ego-planner is a widely adopted grid-map-based planning framework, 
we enlarge its map size to equip it with global path search ability to some extent.
Although cannot run in real-time, FAR-Planner is regarded as the first 2D V-graph based work that is compatible to 3D navigation by using multi-layer polygons.
We selected the garage and the factory for system testing, and travel distance and time as evaluation metrics.

As shown in Table \ref{tab:system}, Fig. \ref{fig:garage} and \ref{fig:factory}, the A*-based system travels significantly longer despite its optimality, because encountering a dead end causes extensive node expansion and slows the search. It performs better in the factory, where A*'s heuristic aids graph search in a well-connected space. For Far-Planner, the system's travel distance and time increase as it waits for graph updates. In contrast, our proposed method continuously updates graphs and performs efficient path searches, ensuring safe and smooth navigation.


\begin{table}[b]
\centering
\caption{System-level Comparison}
\label{tab:system}
\resizebox{0.9\linewidth}{!}{%
\begin{tabular}{|c|cc|cc|}
\hline
Env. & \multicolumn{2}{c|}{Garage}                                                                          & \multicolumn{2}{c|}{Factory}                                                                         \\ \hline
Metrics     & \multicolumn{1}{c|}{\begin{tabular}[c]{@{}c@{}}Travel \\ Dis. (m)\end{tabular}} & Time (s)       & \multicolumn{1}{c|}{\begin{tabular}[c]{@{}c@{}}Travel \\ Dis. (m)\end{tabular}} & Time (s)       \\ \hline
A*          & \multicolumn{1}{c|}{703.8}                                                          & 334.7          & \multicolumn{1}{c|}{1289.6}                                                         & 682.4          \\ \hline
FAR         & \multicolumn{1}{c|}{648.4}                                                          & 341.7          & \multicolumn{1}{c|}{-}                                                              & -              \\ \hline
Ours        & \multicolumn{1}{c|}{\textbf{529.6}}                                                 & \textbf{195.7} & \multicolumn{1}{c|}{\textbf{973.5}}                                                 & \textbf{483.9} \\ \hline
\end{tabular}%
}
\end{table}



\subsection{Real-world Experiment}

We conducted real-world experiments to demonstrate the robustness and adaptability of our system. As illustrated in Fig. \ref{fig:Banner} and Table. \ref{tab:rw_exp}, our algorithm shows robust performance when navigating through the large-scale complex unknown environments with a dead-end. 
The system consistently achieves real-time performance in real-world settings using onboard computing.

\begin{table}[]
\centering
\caption{Real-world Experiment Metrics}
\label{tab:rw_exp}
\resizebox{\linewidth}{!}{%
\begin{tabular}{|c|c|c|c|}
\hline
\begin{tabular}[c]{@{}c@{}}Graph \\ Update (s)\end{tabular} & \begin{tabular}[c]{@{}c@{}}Path \\ Search (s)\end{tabular} & Travel Time (s) & Traj. Len. (m) \\ \hline
0.127                                                       & 0.0053                                                     & 194           & 100.8          \\ \hline
\end{tabular}%
}
\end{table}



\bibliographystyle{IEEEtran}

\bibliography{references}


\end{document}